\documentclass[11pt]{article}
\usepackage{acl2015}
\usepackage{times}
\usepackage{url}
\usepackage{latexsym}
\usepackage{graphicx}

\usepackage{booktabs}
\usepackage{amssymb}
\usepackage{amsfonts}
\usepackage{multicol}
\usepackage{multirow}

\usepackage[colorinlistoftodos]{todonotes}

\title{Egyptian Arabic to English\\
Statistical Machine Translation System\\
for NIST OpenMT'2015}

\author{Hassan Sajjad, Nadir Durrani, Francisco Guzman, Preslav Nakov,\\
\bf Ahmed Abdelali, Stephan Vogel,\\
\bf $\dagger$Wael Salloum, $\dagger$Ahmed El Kholy, $\ddagger$Nizar Habash\\
Qatar Computing Research Institute\\
$\dagger$Columbia University, $\ddagger$New York University Abu Dhabi} 

\date{}

\begin{document}
\maketitle
\begin{abstract}

The paper describes the Egyptian Arabic-to-English statistical machine translation (SMT) system that the QCRI-Columbia-NYUAD (QCN) group submitted to the NIST OpenMT'2015 competition.
The competition focused on informal dialectal Arabic, as used in SMS, chat, and speech.
Thus, our efforts focused on processing and standardizing Arabic, e.g., using tools such as
\emph{3arrib} and \emph{MADAMIRA}. We further trained a phrase-based SMT system using
state-of-the-art features and components such as operation sequence model,
class-based language model, sparse features, neural network joint model, genre-based hierarchically-interpolated language model,
unsupervised transliteration mining,
phrase-table merging, and 
hypothesis combination.
Our system ranked second on all three genres.
\end{abstract}

\section{Introduction}
\label{sec:intro}

We describe the Egyptian Arabic-to-English statistical machine translation (SMT) system of the QCN team for the NIST OpenMT'2015 evaluation campaign. The \underline{QCN} team included the \underline{Q}atar Computing Research Institute, \underline{C}olumbia University, and \underline{N}ew York University in Abu Dhabi.
%

The OpenMT 2015 translation task asked participants to build systems that can translate Egyptian Arabic  from three different genres (SMS, chat, and speech) into English. 
The challenges presented by this multigenre task were many, ranging from scarcity of parallel data for training, to noisy source text and heterogeneity of the references. For example, large portions of the provided data consisted of romanized Egyptian Arabic (aka \emph{Arabizi})
rather than using the Arabic script. 

Therefore, several preprocessing steps were needed in order to clean the data before building our SMT system.
First, we converted all Arabizi to Arabic script; we then normalized it, trying to make it more like MSA. We also morphologically segmented long Arabic words using segmentation schemes that are standard for MSA but harder to do for dialectal Arabic. We used a statistical phrase-based MT system -- Moses~\cite{Moses:2007}. We experimented with different data processing schemes and different SMT system settings to achieve better translation quality.
Here are the major settings: Egyptian Arabic segmentation (ATB, S2, D3), Egyptian Arabic to MSA conversion,
sparse features,
class-based models,
neural network joint language model,
hierarchically-interpolated language model,
unsupervised transliteration mining,
domain adaptation,
and data selection for tuning.


It is worth mentioning that given the above-mentioned challenges, preprocessing by itself yielded the largest gains. The Egyptian Arabic segmentation gave us an improvement of up to 3 BLEU points. The hierarchically-interpolated language model added 1 extra BLEU point. The sparse features, class-based models, and neural network joint language models further improved translation quality by 0.66, 0.70 and 0.43 BLEU points absolute, respectively. 
In the next sections, we discuss in detail the different settings and decisions we made when preparing our submission.

The remainder of the paper is organized as follows:
Section~\ref{sec:preprocessing} explains the data preprocessing techniques and tools we tried,
Section~\ref{sec:system} explains in detail the non-standard features and components we tried in our SMT experiments,
and
Section~\ref{sec:final} describes the actual system we submitted for the competition.
Finally, Section~\ref{sec:conclusion} concludes with a summary of the main points.


\section{Data Preprocessing}
\label{sec:preprocessing}

\begin{table*}[tbh]
\centering
\small
\begin{tabular}{l | r r r r r r}
\toprule
    & \multicolumn{2}{c}{SMS} & \multicolumn{2}{c}{CHT} & \multicolumn{2}{c}{CTS} \\
    & Test & TestG & Test & TestG & Test & TestG\\
\midrule
No segmentation & 21.02 & 21.64 & 20.27 & 22.34 & 20.60 & 23.36 \\
D3  & 23.68 & 23.41 & 23.22 & 25.97 & 21.72 & 24.89 \\
S2  & 23.62 & 23.66 & 22.82 & 25.41 & 21.61 & 24.67 \\
ATB & 23.57 & 23.50 & 22.82 & 26.01 & 21.68 & 24.83 \\
\bottomrule
\end{tabular}
\caption{Comparing different Arabic word segmentation schemes.}
\label{tab:segmentation}
\end{table*}

The NIST dataset contained text from three different genres: short text messages (SMS), chat (CHT), and transcribed conversational speech (CTS). 
We tackled each of the three genres separately, i.e., we built genre-specific systems. Furthermore, we split the provided training data for each genre into separate training and development sets by reserving approximately 3,000 sentences (the number of sentences is approximate as we did the splitting at the document level) from each set for development, and we used the rest for training.

For evaluation, the organizers provided two additional datasets: (\emph{i}) an official devtest dataset (\emph{Test}), and (\emph{ii}) a small gold dataset (\emph{TestG}), which is a subset of \emph{Test}.
The genres of the datasets were specified, and thus there was no need to train a system for automatic genre identification.

\subsection{Modern Standard Arabic Preprocessing}

For training purposes, some datasets for Modern Standard Arabic (MSA) were  provided in addition to the Egyptian Arabic data (SMS, CHT, CTS) described above. These datasets consisted mostly of newswire text (i.e., a different genre), and thus we processed them using a standard MSA tool: MADAMIRA in MSA mode \cite{pasha2014madamira}. 
The MSA data was used to build a second phrase table to be then combined with some of the genre-specific phrase tables.
We planned to match the same morphological tokenization schemes used for Egyptian Arabic, but at the end, we only used ATB segmentation for MSA. For more information on Arabic morphology challenges and tokenization schemes, see  \newcite{Habash-ANLP:2010}.

\subsection{Genre-Specific Preprocessing}

The CTS data contained speech tags, markers for spelling corrections, etc., which we removed before training. 
The Arabic side of the SMS and CHT data had elongations and spelling mistakes. Therefore, we normalized the elongations and we standardized the spelling variations.

Parts of the SMS and CHT data contained romanized Arabic text (aka \emph{Arabizi}), which is Arabic written using the Roman alphabet. In order to homogenize the input data, we converted the Arabizi into the standard Arabic script (utf8).

Given the conversational nature of SMS and chat, a further complication with converting romanized text into the Arabic script was that the text was usually affected by code-switching into English; therefore, it was important for us to identify the English words and not to try to convert them to Arabic. We solved these issues using the {3ARRIB} tool \cite{albadrashiny-EtAl:2014}.

On the English side of the data, there were multiple translation options, where typically the first option was the {\it intended meaning} and the second one the {\it literal meaning} of the Arabic text. We chose to use the {\it intended meaning} only.

Finally, in all processing, for both Arabic and English, we made sure the emoticons were not affected by the tokenization or the translation.

\begin{table*}
\centering
\small
\begin{tabular}{l | r r r r r r}
\toprule
    & \multicolumn{2}{c}{SMS} & \multicolumn{2}{c}{CHT} & \multicolumn{2}{c}{CTS} \\
    & Test & TestG & Test & TestG & Test & TestG\\
\midrule
Original Egyptian, no segmentation & 21.02 & 21.64 & 20.27 & 22.34 & 20.60 & 23.36 \\
Egyptian adapted to MSA, no segmentation & 21.54 & 21.82 & 20.70 & 22.77 & 21.30 & 23.81 \\
Egyptian adapted to MSA, then ATB-segmented & 21.32 & 21.06 & 21.55 & 23.70 & 21.73 & 24.30 \\
\bottomrule
\end{tabular}
\caption{Experiments in adapting Egyptian to look like MSA.}
\label{tab:msa}
\end{table*}

\subsection{Egyptian Arabic Segmentation}

A major issue when training an SMT system for the present edition of the task is the small size of the provided SMS, CHT and CTS datasets, which means that data sparseness is a severe problem. One common way to reduce it, at least on the Arabic side, where it is more severe, is to segment the Arabic words into multiple tokens, e.g., by separating the main word from the attached conjunctions, pronouns, articles, etc. 
Since these are separate words in English, such a segmentation not only reduces sparseness, but also yields improved word mapping to English, thus ultimately helping word alignments and translation model estimation for SMT.
The value of Arabic tokenization for SMT, especially under low resource conditions, has been demonstrated by a number of researchers in the past, e.g.,
 \cite{badr:acl08,Kholy:Habash:2012,MADA-ARZ:2013,Almannai:2014an}. 


We experimented with common segmentation schemes such as D3, S2 and ATB \cite{badr:acl08,Habash-ANLP:2010}. For tokenization, we used MADAMIRA \cite{pasha2014madamira}, a fast and efficient implementation of MADA for MSA \cite{Habash:2005:ATP:1219840.1219911,Habash:2009}, and MADA-ARZ, a version of MADA  for Egyptian Arabic \cite{MADA-ARZ:2013}. Table~\ref{tab:segmentation} compares the results of using different segmentation schemes including no segmentation.
We see gains of up to 3 BLEU points absolute on \emph{Test} when using segmentation compared 
to no segmentation.  However, the differences between the various schemes (ATB, S2 and D3) are small.


\subsection{Egyptian to MSA Conversion}

Another way to reduce data sparseness is by using additional out-of-domain data, e.g., newswire; this means a double domain shift: (\emph{i})~from an informal text genre to newswire, and also  (\emph{ii})~from dialectal Arabic to MSA.
While it is hard to do anything about the domain shift, the dialectal shift is somewhat easier to address. Changes between dialects are often systematic and many of the differences are at the level of individual words.

Previous work has shown that converting Egyptian to MSA makes it easier to use MSA resources for translating dialectal Arabic
\cite{mohamed-mohit-oflazer:2012:ACL2012short,SalloumHabash:2011,Zbib-etal:2012:NAACL,SalloumHabash:2013,Sajjad:2013ac,durrani-EtAl:2014:CICLING}.
So, we experimented with converting Egyptian to MSA using an in-house tool \cite{Sajjad:2013ac}, which performs character-level transformations for each Egyptian word in isolation to generate an MSA version thereof. We then trained an SMT system on this converted data. 

The results are shown in Table~\ref{tab:msa}. We can see that converting Egyptian to MSA yields improvements that are systematic across the three genres and also across the two test datasets. However, this improvement is not very large and ranges from 0.18 to 0.70 BLEU points absolute.

Further segmenting the MSA-like Egyptian using MADA ATB yielded very mixed results: in some cases, it added 0.93 BLEU points absolute, but in other there was a drop of 0.76.
This could be because of the highly dialectal nature of the NIST data, which differs from MSA in lexical choice, which our tool cannot handle. Our Egyptian to MSA tool works at the character level and converts to MSA only those Egyptian words that are different at the character level.

In general, full conversion of dialectal Arabic to MSA would require not only word-level transformations but also phrase-level ones \cite{wang-nakov-ng:2012:EMNLP-CoNLL}, while taking context into account \cite{Nakov:Tiedemann:2012}, and also modeling morphological phenomena \cite{nakov-ng:2011:ACL-HLT2011}. There are also potential gains from a smarter character alignment models \cite{tiedemann-nakov:2013:RANLP-2013}, or even from using a specialized decoder \cite{wang-ng:2013:NAACL-HLT}. Ultimately, the real benefit is when combining the adapted version of the smaller dialect with a large dataset in the bigger dialect \cite{Nakov:Ng:2012}, which we will do below.

\section{Translation System Characteristics}
\label{sec:system}

We started our experiments from a strong baseline system, which was originally designed for MSA to English translation \cite{sajjad-etal:iwslt13}. We then extended it with some additional models and features \cite{Durrani-uedinwmt14}.
Most notably, we used minimum Bayes risk decoding (MBR) \cite{kumar-byrne:2004:HLTNAACL}, monotone-at-punctuation reordering, dropping of out-of-vocabulary words, operation sequence model for reordering (OSM) \cite{durrani-schmid-fraser:2011:ACL-HLT2011,durrani-EtAl:2013:ACL},
a smoothed BLEU+1 version of PRO for parameter tuning \cite{nakov-guzman-vogel:2012:PAPERS}, etc.

Given this baseline system, we experimented with several further extensions, which we will describe below. Some of them were eventually included in our final submission. 



\begin{table*}
\centering
\small
\begin{tabular}{l | r r r r r r}
\toprule
    & \multicolumn{2}{c}{SMS} & \multicolumn{2}{c}{CHT} & \multicolumn{2}{c}{CTS} \\
    & Test & TestG & Test & TestG & Test & TestG\\
\midrule
Baseline using concatenated language model & 24.19 & 24.00 & 23.34 & 25.89 & 22.75 & 25.09\\
System using interpolated language model   & \bf{25.20} & \bf{25.04} & \bf{23.48} & \bf{26.16} & \bf{23.01} & \bf{25.67}\\
\bottomrule
\end{tabular}
\caption{Experiments with interpolated language models of each genre in comparison to the baseline language model built on the concatenation of the English side of SMS, CHT and CTS.}
\label{tab:lm}
\end{table*}

\begin{table*}
\centering
\small
\begin{tabular}{l | r r r r r r}
\toprule
    & \multicolumn{2}{c}{SMS} & \multicolumn{2}{c}{CHT} & \multicolumn{2}{c}{CTS} \\
    & Test & TestG & Test & TestG & Test & TestG\\
\midrule
Baseline using concatenated language model & 24.19 & 24.00 & 23.34 & 25.89 & 22.75 & 25.09\\
System using interpolated language model   & \bf{25.20} & \bf{25.04} & \bf{23.48} & \bf{26.16} & \bf{23.01} & \bf{25.67}\\
\bottomrule
\end{tabular}
\caption{Experiments with interpolated language models of each genre in comparison to the baseline language model built on the concatenation of the English side of SMS, CHT and CTS.}
\label{tab:lm}
\end{table*}

\begin{table*}
\centering
\small
\begin{tabular}{l | r r r r r r}
\toprule
    & \multicolumn{2}{c}{SMS} & \multicolumn{2}{c}{CHT} & \multicolumn{2}{c}{CTS} \\
    & Test & TestG & Test & TestG & Test & TestG\\
\midrule
Baseline & 24.58 & 24.82 & 23.36 & 26.11 & 22.64 & 24.95\\
+ sparse features & 24.54 & 25.36 & \bf{24.02} & \bf{27.11} & 21.61 & 24.08\\
\bottomrule
\end{tabular}
\caption{Experiments with sparse features.}
\label{tab:sparseFeatures}
\end{table*}

\subsection{Genre-Based Hierarchically-Interpolated Language Model}

First, we experimented with building a hierarchically-interpolated language model for each genre (i.e., CTS, SMS, and CHT).
We tuned each model to minimize the perplexity on a held-out set for that target genre.

We examined the text resources that were available for training an English language model, and we split them into six groups:
(1)~Egyptian-source (the target sides of the CTS, CHT and CTS training bi-texts),
(2)~MSA GALE News (GALE P3 \{R1,R2\},P4\{R1,2,3\}),
(3)~Chinese GALE (GALE P2 \{BC,BC,BL,NG\}),
(4)~MSA NEWS (news-etirr, news-par, news-trans, ISI),
(5)~MSA GALE non-news (GALE P1 \{BLOG\} P2 \{BC1, BC2, WEB\}), and
(6)~Gigaword v5, split into four subgroups by year \cite{guzman-EtAl:2012:WMT} (1994-1997, 1998-2001, 2002-2005, 2006-2010).

For each such group,
(\emph{i}) we built an individual 5-gram language model with Kneser-Ney smoothing for each member of the group, and
(\emph{ii}) we interpolated these language models into a single language model, minimizing the perplexity for the target genre.
Then, (\emph{iii})  we performed a second (hierarchical) interpolation, this time combining the resulting six group language models,  
again minimizing the perplexity for the target genre.
We used the SRILM toolkit \cite{Stolcke:2002}  to build the language models. 


Table \ref{tab:lm} shows the results of using interpolated language model for each domain in comparison with using a language model built on the concatenation of the English side of the SMS, CHT and CTS corpora. The interpolated language model consistently improved all genres with a maximum improvement of up to 1 BLEU points on \emph{Test}.

\subsection{Translation Model with Sparse Features}

The next thing we experimented with were sparse features
\cite{chiang-knight-wang:2009:NAACLHLT09}, which are a recent addition to the Moses SMT toolkit. In particular, we used target and source word insertion features: (\emph{i}) top 50, and (\emph{ii}) all. The latter worked better, and thus we only show results for it.

The results for \emph{all} are shown in Table~\ref{tab:sparseFeatures},
where we can see that sparse features only helped for CHT, while they were harmful for CTS, and yielded mixed results for SMS.
Thus, in our final system, we only used them for CHT.

\subsection{Class-Based Language Models}

Next, we experimented with using automatic word clusters, which we computed on the source and on the target sides of the training bi-text using \emph{mkcls}. We also experimented with OSM models \cite{durrani-EtAl:2013:NAACL} 
over cluster IDs \cite{Durrani-osm-coling14,Bisazza-coling14}. Normally, the lexically-driven OSM model falls back to context sizes of 2-3 operations due to data sparseness, but learning operation sequences over cluster IDs enabled us to learn richer translation and reordering patterns that can generalize better in sparse conditions.

Table~\ref{tab:brownCluster} shows the experimental results when adding a target language model and an OSM model over cluster IDs. We can see that these class-based models yielded consistent improvements in all cases. We also tried using \emph{word2vec} \cite{mikolov-yih-zweig:2013:NAACL-HLT} for clustering, but the results did not improve any further and they were occasionally worse than those with \emph{mkcls}. We tried both with 50 and 500 classes, but using additional classes did not help.

\begin{table*}[tbh]
\centering
\small
\begin{tabular}{l | r r r r r r}
\toprule
    & \multicolumn{2}{c}{SMS} & \multicolumn{2}{c}{CHT} & \multicolumn{2}{c}{CTS} \\
    & Test & TestG & Test & TestG & Test & TestG\\
\midrule
Baseline & 24.22 & 24.33 & 23.02 & 25.60 & 21.93 & 24.88\\
+ class-based models & \bf{24.63} & \bf{25.16} & 23.18 & \bf{26.30} & 22.20 & 25.04\\
\bottomrule
\end{tabular}
\caption{Experiments with class-based language models.}
\label{tab:brownCluster}
\end{table*}

\begin{table*}[tbh]
\centering
\small
\begin{tabular}{l | r r r r r r}
\toprule
    & \multicolumn{2}{r}{SMS} & \multicolumn{2}{r}{CHT} & \multicolumn{2}{r}{CTS} \\
    & Test & TestG & Test & TestG & Test & TestG\\
\midrule
Baseline & 24.58 & 24.33 & 24.02 & 27.11 & 22.64 & 24.95\\
+ NNJM Model & \bf{25.01} & 25.72 & 24.24 & 27.41 & 22.68 & 25.21\\
\bottomrule
\end{tabular}
\caption{Experiments with a neural network joint language model.}
\label{tab:nnjm}
\end{table*}

\subsection{Unsupervised Transliteration Models}

A consequence of data sparseness is that at test time, the SMT system would see many unknown or out-of-vocabulary (OOV) words. One way to cope with them is to just pass them through untranslated. This works somewhat fine with newswire text and for languages with (roughly) the same alphabet, e.g., English and Spanish, as many OOVs are likely to be named entities (persons, locations, organizations), and are thus likely to be preserved in translation.

However, for languages with different scripts, such as Arabic and English, passing through is not a good idea, especially when translating into English as words in Arabic script do not naturally appear in English. In that case, it is much safer just to drop the OOVs, which is best done at decoding time; this was indeed our baseline strategy.

A better way is to transliterate OOV words either during decoding or in a post-processing step \cite{sajjad2013qcri}. 
We also 
experimented with this approach.
For this purpose, we built an unsupervised transliteration model \cite{durrani-EtAl:2014:EACL} based on EM as proposed in \cite{sajjad:acl11}. Unfortunately, it did not help much, probably because in these informal genres the OOVs are rarely named entities; they are real words, which need actual translation, not transliteration.


\subsection{Neural Network Joint Language Model}

Recently, neural networks have come back from oblivion with the promise to revolutionize NLP. Major performance gains have already been demonstrated for speech recognition, and there have been successful applications to semantics. Most importantly for us, last year, very sizable performance gains were also reported for SMT using a neural joint language model or NNJM \cite{devlin2014fast}.

We tried the Moses implementation of JNLM using the settings described in \cite{birch-etal:2014:IWSLT}.
While we managed to achieve consistent improvements for all three genres and for both test sets, as Table~\ref{tab:nnjm} shows, these gains are modest, 0.04--0.57 BLEU points absolute, which is far from what was reported in \cite{devlin2014fast}. It is unclear what the reasons are, but it could have to do with the small size of our training bitexts and the informal genres we are dealing with.

\subsection{Domain Adaptation}

We experimented with different techniques for domain adaptation, trying to combine bi-texts from different genres, e.g., our Egyptian SMS, CHT, and CTS, but also MSA newswire.

First, we experimented with concatenating our SMS, CHT and CTS bitexts for training, but then using genre-specific tuning sets; this did not work as well as some other alternatives.
Next, we experimented with building separate phrase tables, one in-domain and one out-of-domain, and then (a)~using phrase table backoff, or (b)~merging phrase tables and reordering tables as in \cite{nakov:2008:WMT,nakov-ng:2009:EMNLP,sajjad-etal:iwslt13}.

The results of our domain adaptation experiments when testing on SMS as the target genre are shown in Table~\ref{tab:sms_adpt}. 
Note that the results on \emph{Test} and on \emph{TestG} differ a lot, and thus we focus on \emph{Test} as it is much larger. We can see that the best way to combine SMS, CHT and CTS is simply to concatenate them, which yields +2.5 BLEU points of improvement absolute over training on SMS data only. Further small gains can be achieved by merging the resulting SMS+CHT+CTS phrase table with a phrase table trained on MSA, where the two tables are merged using extra indicator features as described in \cite{nakov:2008:WMT}.

\begin{table}[tbh]
\centering
\small
\begin{tabular}{l | r r}
\toprule
Training data & Test & TestG \\
\midrule
SMS & 21.30 & 21.99 \\
CAT(SMS, CHT, CTS) & \bf{23.78} & 23.20 \\
SMS, Backoff(CHT,CTS) & 22.55 & 23.00 \\
CAT(SMS, CHT), Backoff(CTS) & 22.54 & 23.20 \\
MergePT(CAT(SMS, CHT), CTS) & 23.69 & \bf{24.40} \\
\midrule
CAT(SMS, CHT, CTS), Backoff(MSA) & 23.70 & \bf{23.64} \\
MergePT(CAT(SMS, CHT, CTS), MSA) & \bf{23.83} & 23.60 \\
\bottomrule
\end{tabular}
\caption{Experiments with different training data combinations (bitext concatenations, phrase table merging, and backoff), testing on SMS as the target genre.}
\label{tab:sms_adpt}
\end{table}

\subsection{Tuning}

Our phrase-based SMT system combines different features in a log-linear model. We tune the weights for the individual features of that model by optimizing BLEU \cite{Papineni:Roukos:Ward:Zhu:2002}
on a tuning dataset from the same genre as that in the test. We use PRO \cite{Hopkins2011}, but with smoothed BLEU+1 as proposed in \cite{nakov-guzman-vogel:2012:PAPERS}. We allowed the optimizer to run for up to 25 iterations, and to extract 1000-best lists on each iteration.

The choice of tuning set has been shown to have a huge impact on the quality of the learned parameters \cite{nakov:2013:parameter}. In particular, PRO is very sensitive to length, which can result in pathological translations in some circumstances \cite{nakov2013tale}. 

Given that there were no official development sets for this year, we synthesized datasets that are specific for CTS and SMS, based on sentence-length as a selection criterion.\footnote{The CHT sentences were generally of reasonable length, and thus we did not apply filtering for them.} This is crucial to remove potentially noisy data. It can also help to speed up the tuning process.

In order to achieve this, we filtered all sentence pairs for which either the source or the target sentences were shorter than 4 words or longer than 25 words. The cut-offs were determined empirically by analyzing the kernel density estimations (KDE $\alpha=0.3$).

Table~\ref{tab:tuneclean} compares tuning on an unfiltered and on a filtered tuning set. For SMS, using a filtered tuning set yields mixed results: we see a drop in BLEU on \emph{Test} and a gain on \emph{TestG}. For CTS, filtering helped both on \emph{Test} and on \emph{TextG}, with a gain on the former of +0.63.

\begin{table}
\centering
\small
\begin{tabular}{l | r r}
\toprule
Genre & Test & TestG \\
\midrule
SMS - unfiltered tune & 23.57 & 23.56 \\
SMS - filtered tune & 23.35 & 24.36 \\
\midrule
CTS - unfiltered tune & 22.07 & 25.10 \\
CTS - filtered tune & 22.70 & 25.22 \\
\bottomrule
\end{tabular}
\caption{Results using a length-filtered vs. an unfiltered dataset for tuning.}
\label{tab:tuneclean}
\end{table}

\section{Final Submission and Output Combination}
\label{sec:final}

We recombined hypotheses produced (a)~by our best individual systems and (b)~by other systems that are both relatively strong and can contribute diversity,
e.g., using a different word segmentation scheme. For this purpose, we used the Multi-Engine MT system, or MEMT, \cite{Heafield-wmt09}, which has been proven effective in such a setup and has contributed to achieving state-of-the-art results in a related competition in the past \cite{sajjad-etal:iwslt13}.

The results are shown in Table~\ref{tab:syscomb}. We can see that using output combination yields notable improvement for SMS and CHT. However, for CTS, BLEU dropped by 0.45 points on \emph{Test}. Thus, we submitted as a primary system the output combination for SMS and CHT, but our best individual system for CTS (which uses D3 segmentation).



\begin{table*}
\centering
\small
\begin{tabular}{l| r r r r r r}
\toprule
	& \multicolumn{2}{c}{SMS} & \multicolumn{2}{c}{CHT} & \multicolumn{2}{c}{CTS} \\
 	& Test & TestG & Test & TestG & Test & TestG\\
\midrule
Best system with D3 segmentation & 25.28 & 26.05 & 23.87 & 27.07 & {\bf 23.34} & {\bf 26.05} \\
Best system with S2 segmentation & 24.93 & 25.61 & 24.09 & 27.01 & 22.11 & 24.50 \\
Best system with ATB segmentation & 25.13 & 25.80 & 24.24 & 27.41 & 22.83 & 25.56 \\
System with ATB segmentation + MSA phrase table as a backoff & 25.20 & 25.04 & 23.48 & 26.16 & 23.01 & 25.67 \\
\midrule
Output combination & {\bf 26.13} & {\bf 26.79} & {\bf 24.86} & {\bf 27.95} & 22.89 & 25.88 \\
\bottomrule
\end{tabular}
\caption{Results for system recombination.}
\label{tab:syscomb}
\end{table*}

\section{Conclusion}
\label{sec:conclusion}

We presented the Egyptian Arabic-to-English SMT system of the QCN team for the NIST OpenMT'2015 evaluation campaign. The system was ranked second in the competition on all three genres: SMS, chat, and speech.

Given the informal dialectal nature of these genres, we benefited from careful pre-processing, cleaning, and normalization, which yielded an improvement of up to 3 BLEU points over 
a strong baseline.

We further added a number of extra advanced features, which yielded 2.5 more BLEU points of absolute improvement on top of that due to pre-processing.
In particular, sparse features contributed 0.7 BLEU points for CHT, class-based models added 0.7 and 0.6 BLEU points for CHT and SMS, respectively, and NNJM yielded gains of up to 0.4 BLEU points absolute.





\bibliographystyle{acl}
\bibliography{acl2015}

\end{document}